\documentclass[letterpaper, 10 pt, conference]{ieeeconf}  

\IEEEoverridecommandlockouts                              

\usepackage{graphics} 
\usepackage{epsfig} 
\usepackage{mathptmx} 
\usepackage{times} 
\usepackage{amsmath} 
\usepackage{amssymb}  
\usepackage[ruled,vlined]{algorithm2e}
\let\labelindent\relax
\usepackage{enumitem}

\title{\LARGE \bf
Bike Sharing Demand Prediction based on Knowledge Sharing across Modes: A Graph-based Deep Learning Approach
}

\author{Yuebing Liang$^{1}$, Guan Huang$^{2}$ and Zhan Zhao$^{3}$
\thanks{$^{1}$Yuebing Liang is with Department of Urban Planning and Design, The University of Hong Kong, Hong Kong, China {\tt\small yuebingliang@connect.hku.hk}}%
\thanks{$^{2}$Guan Huang is with Department of Urban Planning and Design, The University of Hong Kong, Hong Kong, China {\tt\small guanhuang@connect.hku.hk}}%
\thanks{$^{3}$Zhan Zhao is with Department of Urban Planning and Design, The University of Hong Kong, Hong Kong, China {\tt\small zhanzhao@hku.hk}}%
}

\begin{document}

\maketitle
\thispagestyle{empty}
\pagestyle{empty}

\begin{abstract}

Bike sharing is an increasingly popular part of urban transportation systems. Accurate demand prediction is the key to support timely re-balancing and ensure service efficiency. Most existing models of bike-sharing demand prediction are solely based on its own historical demand variation, essentially regarding bike sharing as a closed system and neglecting the interaction between different transport modes. This is particularly important because bike sharing is often used to complement travel through other modes (e.g., public transit). Despite some recent efforts, there is no existing method capable of leveraging spatiotemporal information from multiple modes with heterogeneous spatial units. To address this research gap, this study proposes a graph-based deep learning approach for bike sharing demand prediction (B-MRGNN) with multimodal historical data as input. The spatial dependencies across modes are encoded with multiple intra- and inter-modal graphs. A multi-relational graph neural network (MRGNN) is introduced to capture correlations between spatial units across modes, such as bike sharing stations, subway stations, or ride-hailing zones. 
Extensive experiments are conducted using real-world bike sharing, subway and ride-hailing data from New York City, and the results demonstrate the superior performance of our proposed approach compared to existing methods. 

\end{abstract}

\begin{keywords}
bike sharing, demand prediction, inter-modal relationships, graph neural networks, deep learning
\end{keywords}

\section{INTRODUCTION}\label{intro}

In the past decade, bike sharing has emerged as a sustainable, convenient, and generally affordable travel mode, and become an integral component of urban transportation systems in cities around the world. 
The trend is likely to continue under the ongoing COVID-19 pandemic, as biking is usually seen as a safer and healthier travel option. Because of its positive effects on the environment, public health and traffic congestion, bike sharing systems (BSS) should be promoted to play a bigger role in urban mobility. Currently, efficient operations of BSS rely on the dynamic rebalancing of bikes between locations to better match the ever-changing demand patterns \cite{liu2016rebalancing}. Therefore, the accurate short-term demand prediction at high spatial resolution is crucial because it is the basis to ensure the availability, efficiency, and user experience of bike sharing service.

Recent years have seen growing interests in short-term demand prediction for intelligent BSS, with a particular focus on deep learning methods, because of their demonstrated effectiveness in extracting the complex and nonlinear knowledge hidden in large-scale mobility data \cite{xu2018station, jin2020dockless}. Despite the success of these methods, they regard bike sharing as a closed system and neglect the potential rich information of the interaction between BSS and other transportation modes. This is especially important to consider for bike sharing, because it is mainly used for short-distance trips or the first-mile/last-mile portion of longer trips. In practice, BSS are often designed as feeders to public transport systems or support multimodal transport connections \cite{zhang2018short}. Extensive prior studies have explored the relationship between BSS usage with public transit, taxi and ride-hailing, and unrevealed significant complementary or competitive relationships subject to trip purpose, trip length, availability, etc \cite{song2020investigating,gu2019measuring}. As a result, the demand for bike sharing will inevitably be influenced by other transportation modes, which should be considered in demand prediction. Incorporating demand information across modes can also help mitigate the data sparsity problem commonly seen in BSS, since bike sharing is rarely one of the primary travel modes in cities.


To incorporate inter-modal demand information for bike sharing demand prediction, recent studies have proposed to relate each BSS station to its adjacent subway stations or bus stops and use their accumulated usage as an additional attribute of BSS stations \cite{zhang2018short, cho2021enhancing}. Another approach is to divide the study area into uniform grids and aggregate multimodal data to the same grid system \cite{ye2019co,wang2022nonlinear}. However, these methods are not always straightforward because of the heterogeneous spatial units of different transportation modes: some are station-based (e.g., subway, station-based BSS), while others are stationless (e.g., ride-hailing, dockless BSS) \cite{liang2021joint}. Instead of arbitrarily aligning multimodal networks, we need a more flexible approach that can learn the inter-modal relationships across heterogeneous spatial units directly from data. Also, the learning should be done with high spatial resolution so that it is useful for BSS operations. For example, dynamic rebalancing for station-based BSS would require good demand prediction at the station level.
  

In this paper, we propose a graph-based deep learning approach for bike sharing demand prediction (B-MRGNN). To integrate multimodal data with diverse spatial units, we encode spatial dependencies across different modes with multiple intra- and inter-modal graphs. To extract cross-mode relationships, we introduce an improved version of the multi-relational spatiotemporal graph neural network from our prior work \cite{liang2021joint}. The proposed model is demonstrated using the Citi Bike data from New York City (NYC), with the subway and ride-hailing data as additional inter-modal demand input. Because of data constraint, we will focus on station-based BSS in this study, but the methodology should be generalizable to stationless (or dockless) BSS as well. The specific contributions of this research are as follows:

\begin{itemize} [noitemsep]
    \item We propose a deep learning model for bike sharing demand prediction based on inter-modal relationships learnt from historical demand data. 
    \item We introduce a multi-relational graph neural network (MRGNN) to model spatial correlations between heterogeneous spatial units across different modes. 
    \item Extensive experiments are conducted based on real-world datasets from NYC, and the results demonstrate the superior performance of our proposed model compared to existing methods.
\end{itemize}



\section{LITERATURE REVIEW}\label{literature}

Traditional methods for bike sharing demand prediction mainly focus on finding the relationships between bike demand and its historical demand and exogenous factors (e.g. weather, land use) 
through regression models, such as auto-regressive integrated moving average (ARIMA) \cite{yoon2012cityride} and linear regression (LR) \cite{rudloff2014modeling}. Later research 
adopts other machine learning models, including random forest \cite{guidon2020expanding} and gradient boosting machines \cite{feng2018hierarchical}. 
In recent years, extensive studies have presented the power of deep learning models to capture the nonlinear and complex relationships for demand prediction tasks. Xu et al. \cite{xu2018station} introduced a recurrent-based model considering the effect of exogenous factors.
Jin et al. \cite{jin2020dockless} applied temporal convolution networks (TCNs) to capture temporal dependencies for bike sharing demand prediction. 
To incorporate spatial information, researchers employed convolutional neural networks (CNNs) along with recurrent layers to capture complex spatiotemporal influences, using artificial grids for demand aggregation \cite{zhou2018predicting,li2021short}. 
Graph neural networks (GNNs) exempt the requirement of artificial segmentation and can capture relationships at the station level, which is more useful for BSS operations. Yu et al. \cite{yu2017spatio} introduced a convolution-based model with Graph convolutional networks (GCNs) to model spatial dependencies and TCNs to model temporal dependencies. A multi-graph learning approach was proposed in \cite{geng2019spatiotemporal} for ride-hailing demand prediction encoding multiple types of spatial dependencies. Wu et al. \cite{wu2019graph} introduced an adaptive adjacency matrix to learn spatial dependencies hidden in data. However, these methods are mode-specific and do not consider inter-modal relationships.

Only a few studies have considered the influence of other modes for bike sharing demand prediction. Zhang et al. \cite{zhang2018short} incorporated the historical demand of public transit and used LSTM for bike sharing demand prediction. Cho et al. \cite{cho2021enhancing} enhanced the accuracy of bike sharing demand during peak hours with public transit usage information using a graph learning approach. These methods assign the accumulated flow of adjacent public transit stations to each BSS station, and do not model the spatiotemporal dependencies among adjacent stations/zones from different modes directly. Recent research has also investigated the co-prediction of bike sharing demand and other transport modes. In \cite{ye2019co} and \cite{wang2022nonlinear}, the demands for taxis and bike sharing are are aggregated to a grid system to enable shareable feature learning, before co-predicted using a convolutional recurrent network. Despite existing relevant works, a model that can directly leverage spatiotemporal knowledge across modes with heterogeneous spatial units is still needed. 

\section{Problem Statement}\label{Problem}
In this section, we introduce some notations in this research and then formulate our problem.  

\textit{Definition 1 (Demand Sequence):} Consider a transport mode $m$ with $N_m$ nodes (i.e. stations/service zones). For each node $i = 1, 2, ..., N_m$, its inflow and outflow demand at time step $t$ is denoted as $x_{m, i}^t\in \mathbb{R}^2$. Next, we represent the demand of all the nodes from mode $m$ at time step $t$ as $X_m^t=\{x_{m,0}^t, x_{m,1}^t…, x_{m,N_m}^t\},X_m^t \in \mathbb{R}^{N_m \times 2}$. Further, we use $X_m^{t-T:t}=\{X_m^{t-T}…,  x_m^{t-1},x_m^{t}\}$ to denote the demand sequence of mode $m$ over time steps $T$.

\textit{Problem (Bike Sharing Demand Prediction):} This research aims to predict the station-level bike sharing demand given historical demand of BSS as well as other modes. Formally, given the historical demand of bike sharing denoted as $X_{b}^{t-T:t}$ and auxiliary modes, i.e., subway and ride-hailing in our case, denoted as $X_{s}^{t-T:t}$ and $X_{h}^{t-T:t}$, the goal is to predict bike sharing demand $X_{b}^{t+1}$ at the next time step : 
\begin{equation}
X_{b}^{t+1}=F(X_{b}^{t-T:t},X_{s}^{t-T:t},X_{h}^{t-T:t}),
\end{equation}
where $F(*)$ is the prediction function to be learned by our proposed model. This formulation can be easily adapted to other demand prediction problems with multimodal historical demand as input.


\section{METHODOLOGY}\label{method}
\begin{figure*}[ht!]
  \centering
  \includegraphics[width=0.9\linewidth]{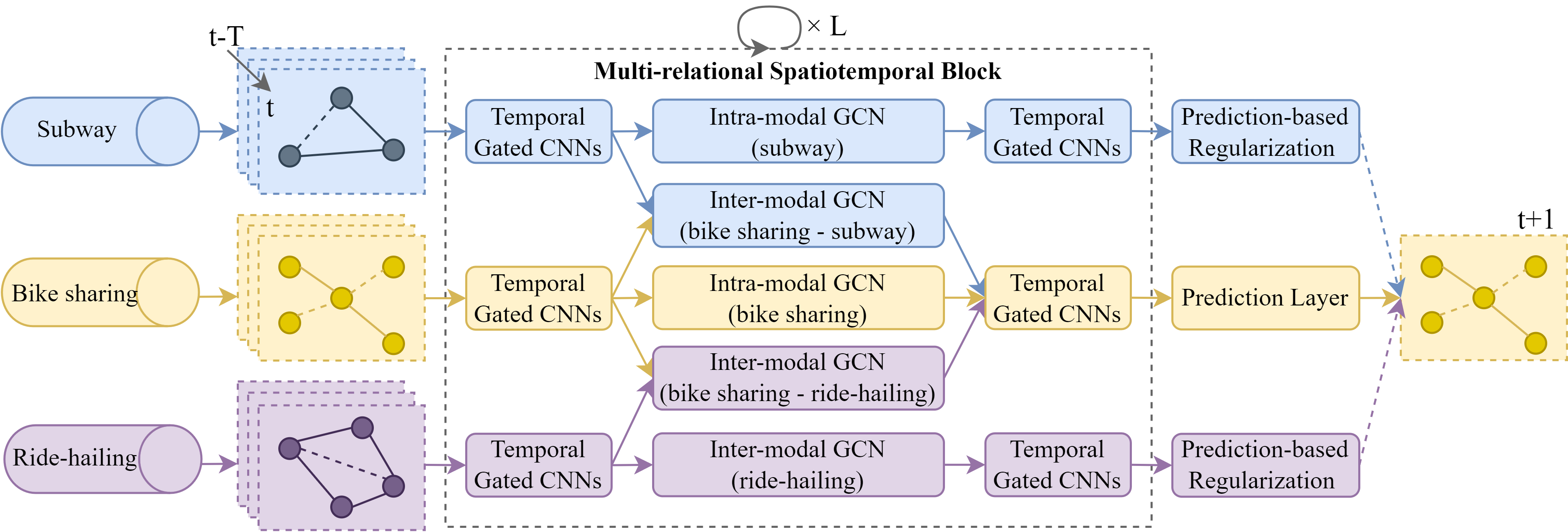}
  \caption{The architecture of B-MRGNN 
  }
  \label{fig:ST-MRGNN}
\end{figure*}

This section presents an improved version of the multi-relational spatiotemporal graph neural network framework \cite{liang2021joint} for station-level bike sharing demand prediction (B-MRGNN) by taking advantage of historical demand sequences of subway and ride-hailing systems. Figure~\ref{fig:ST-MRGNN} displays the overall architecture of our proposed model. It is composed of $L$ multi-relational spatiotemporal blocks (ST-MR blocks) for multimodal representation learning, each comprising TCNs to model temporal patterns and multi-relational graph neural networks (MRGNNs) to model the spatial influence of adjacent subway stations and ride-hailing zones on BSS stations. Based on the learned representation from ST-MR blocks, a prediction layer is used to generate bike sharing demand prediction. We further introduce a prediction-based regularization term that incorporates the demand prediction error for subway and ride-hailing during model training. Details of each module are introduced below.

\subsection{Multi-relational graph neural network}\label{method:MRGNN}
\begin{figure}[ht!]
  \centering
  \includegraphics[width=\linewidth]{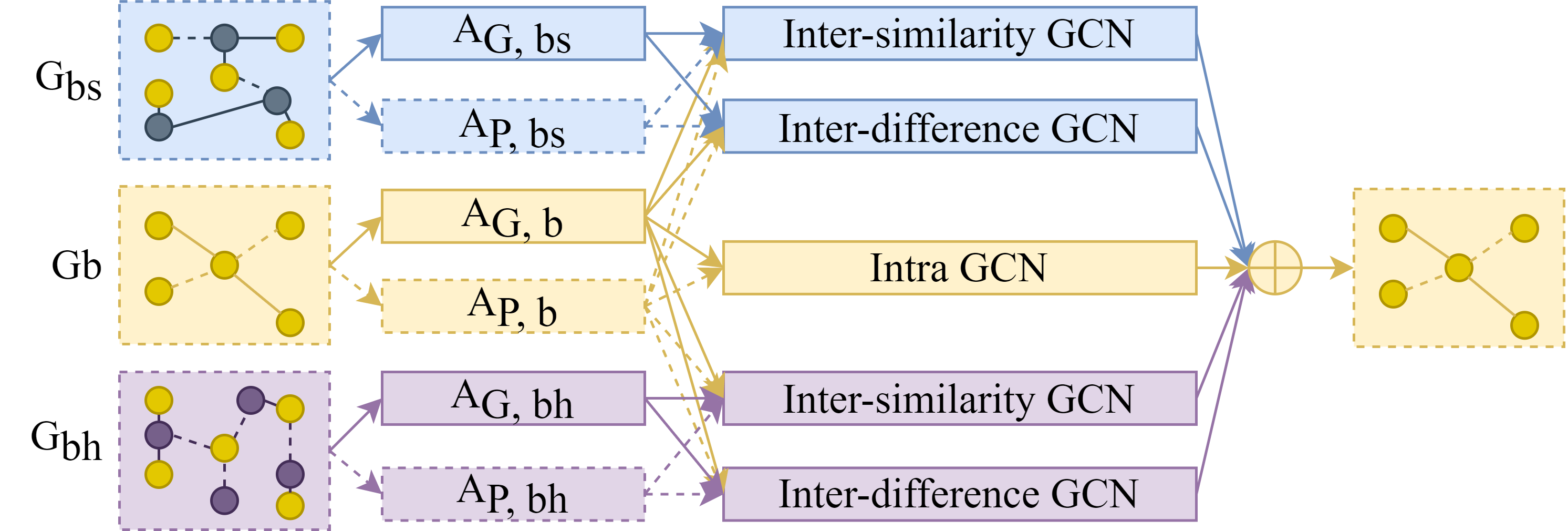}
  \caption{The framework of MRGNN}\label{fig:MRGNN}
\end{figure}

In this subsection, we provide more details on the MRGNN that is used to capture the interactions between heterogeneous spatial units across modes. As shown in Figure \ref{fig:MRGNN}, MRGNN is composed of two major parts: \textit{multi-relational graph construction} to encode cross-mode spatial dependencies and \textit{multi-relational graph convolutions} to capture correlations between nodes through message passing.

\subsubsection{Multi-relational graph construction}\label{method: graph_construct}
As subway and BSS (in our case) are station-based and ride-hailing is stationless, it is difficult to model spatial dependencies on a single homogeneous graph. To encode spatial dependencies within and across modes, we define multiple intra- and inter-modal graphs. Specifically, \textbf{intra-modal graphs} are used to capture spatial correlations among stations/zones of the same mode. Taking bike sharing as an example, its intra-modal graph is defined as $G_{b} = (V_b, A_b)$, where $V_b$ is a set of BSS stations, and $A_b \in \mathbb{R}^{N_b \times N_b}$ is an adjacency matrix representing the spatial dependencies between BSS stations. Similarly, an intra-modal graph is defined for subway and ride-hailing, denoted as $G_{s}$ and $G_{t}$.
\textbf{Inter-modal graphs} are defined to capture the pairwise correlations among stations/zones between the target mode (i.e., bike sharing) and each of the auxiliary modes (i.e., subway and ride-hailing). For example, the inter-modal graph between bike sharing and subway is represented as $G_{bs}=(V_b, V_s, A_{bs})$, where $V_b$ and $V_s$ denote BSS and subway stations and $A_{bs} \in \mathbb{R}^{N_b \times N_s}$ is a weighted matrix indicating the cross-mode dependencies between adjacent BSS and subway stations. Similarly, an inter-modal graph is defined between bike sharing and ride-hailing denoted as $G_{bh}$. 

Prior studies have demonstrated that strong correlations may occur between locations that are either geographically close or semantically similar (i.e., demand patterns) \cite{geng2019spatiotemporal}. To encode both relationships, we define two adjacency matrices for each graph: one for \textbf{geographical proximity} denoted as $A_G$ and, the other for \textbf{semantic similarity} denoted as $A_P$.
The former is computed as a function of distance, while the latter based on demand patterns. The specific definitions of $A_G$ and $A_P$ follow our prior work \cite{liang2021joint}.

\begin{figure}[ht!]
  \centering
  \includegraphics[width=\linewidth]{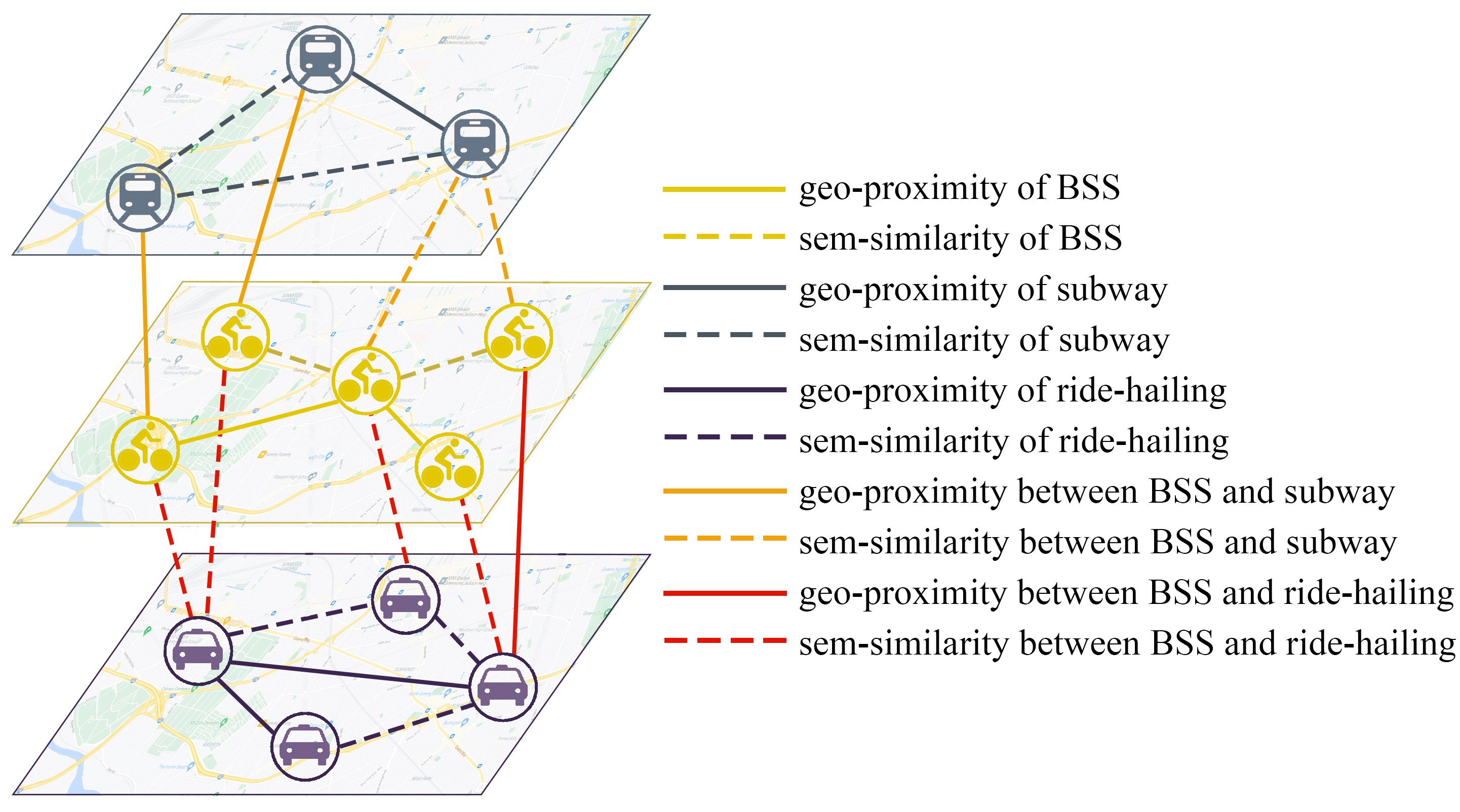}
  \caption{Modeling multimodal spatial dependencies for BSS}\label{fig:graph construct}
\end{figure}

Figure \ref{fig:graph construct} illustrates the constructed multi-relational graph of bike sharing, subway and ride-hailing. A total of (3+2)x2=10 relations is defined to encode spatial dependencies between nodes from different modes, including 3 intra-modal and 2 inter-modal graphs, each with 2 adjacency matrices. This formulation can be easily adapted to encode spatial dependencies across other modes with diverse network structures spatial. 

\subsubsection{Multi-relational graph convolutions}\label{graph_convolution}

Graph convolutions have been an effective way to aggregate information of connected nodes. However, most existing GCNs cannot be applied to the multi-relational graph defined above for two main reasons: first, they cannot process inter-modal graphs with heterogeneous nodes and a non-square adjacency matrix; second, they fail to consider the disparity between different modes, which might result in negative transfer. To tackle these issues, we introduce an inter-modal graph convolution network to aggregate information of connected subway stations and ride-hailing zones for each BSS station, considering both cross-mode similarity and difference. 
Given the inter-modal graphs $G_{bs}=(V_b, V_s, A_{bs})$ and $G_{bh}=(V_b, V_h, A_{bh})$, as well as the input representations of subway and ride-hailing, denoted as $H_s \in \mathbb{R}^{N_s \times c_{in}}$, $H_h \in \mathbb{R}^{N_h \times c_{in}}$ respectively, the \textbf{inter-modal similarity} is modeled as:
\begin{gather}
Z_{bs}^{(c)} =  ReLU({\widetilde{A}_{b}} (\widetilde{A}_{bs}H_s) W_{bs}^{(c)}+l_{bs}^{(c)}),\\
Z_{bh}^{(c)} =  ReLU({\widetilde{A}_{b}} (\widetilde{A}_{bh}H_h) W_{bh}^{(c)}+l_{bh}^{(c)}),
\end{gather}
where $Z_{bs}^{(c)} \in \mathbb{R}^{N_b \times c_{out}}$, $Z_{bh}^{(c)} \in \mathbb{R}^{N_b \times c_{out}}$ are the aggregated features of connected subway stations and ride-hailing zones on BSS stations respectively, $W_{bs}^{(c)}, W_{bh}^{(c)} \in \mathbb{R}^{c_{in} \times c_{out}}$ and $l_{bs}^{(c)}, l_{bh}^{(c)} \in \mathbb{R}^{c_{out}}$ are the learned model parameters, $c_{in}$ and $c_{out}$ represents the input and output vector dimension of each node (i.e. station/zone). ${\widetilde{A}} = \frac{A}{rowsum(A)}$ denotes the normalized adjacency matrix constructed from $A$. 

Our preliminary analysis shows that there is a notable disparity between the demand patterns of different modes: generally, the demand of subway and ride-hailing are more concentrated during rush hours, while bike sharing usage is more active during off-peak period. 
Simply modeling cross-mode similarity can easily lead to negative transfer. Inspired by \cite{zhou2021modeling}, we model \textbf{inter-modal difference} between bike sharing and the two auxiliary modes as:
\begin{gather}
Z_{bs}^{(d)} =  ReLU({\widetilde{A}_{b}}|\widetilde{A}_{bs}H_s - H_b| W_{bs}^{(d)}+l_{bs}^{(d)}),\\
Z_{bh}^{(d)} =  ReLU({\widetilde{A}_{b}}|\widetilde{A}_{bh}H_h - H_b| W_{bh}^{(d)}+l_{bh}^{(d)}),
\end{gather}
where $|\widetilde{A}_{bs}H_s - H_b|$, $|\widetilde{A}_{bh}H_h - H_b|$ represent the information gap between BSS stations and the aggregated information of connected subway stations and ride-hailing zones respectively, $W_{bs}^{(d)}, W_{bh}^{(d)}, l_{bs}^{(d)}, l_{bh}^{(d)}$ are the learned model parameters.

In addition, we apply a standard GCN layer to model pairwise correlations between connected BSS stations. Given the intra-modal graph $G_b$ and the input representations of bike $H_b \in \mathbb{R}^{N_b \times c_{in}}$, the correlations between BSS stations are modeled as:
\begin{equation}
{Z}_{b} =  ReLU({\widetilde{A}_{b}} H_b {W}_{b}^{c}+l_{b}^{c}),
\end{equation}
where ${W}_{b}^{c}, l_{b}^{c}$ are model parameters. In practice, the inter-modal similarity and difference graph convolution layers as well as the intra-modal graph convolution layer can be modeled in parallel using batch matrix multiplication operations. Through the intra- and inter-modal graph convolutions, each BSS station receives multiple feature vectors from geographically adjacent or semantically similar subway stations, ride-hailing zones and other BSS stations. The learned feature vectors from heterogeneous neighborhood nodes are then aggregated using an adding function.   

\subsection{Multi-relational spatiotemporal block} \label{method:represent_learn}
In this subsection, we describe the design of ST-MR blocks, which are stacked to capture correlations between nodes of different modes from both spatial and temporal domains. To capture temporal dependencies, we employ a temporal gated convolution network (TCN) proposed by \cite{yu2017spatio}, due to its high training efficiency and good prediction performance. Briefly, given the historical demand series of a node, TCN models the relationship between each time step and its neighborhoods using a 1-D causal convolution. As introduced above, different modes usually exhibit notably different temporal patterns. Therefore, we apply a separate TCN layer for each mode to capture mode-specific temporal information. The learned features of different modes from TCN layers are then fed into inter-modal graph convolution layers to jointly model spatial and temporal dependencies of BSS stations on connected subway stations and ride-hailing zones. We also apply an intra-modal graph convolution layer on each mode to capture spatiotemporal correlations within modes. 
Following \cite{liang2021joint}, we use an additional TCN layer for each mode after MRGNN layers. 
The input of the second TCN layer is the sum of the output of the first TCN layer and the MRGNN layer, as a residual connection function to speed up model training. To stablize model parameters, we employ a layer normalization function at the end of each ST-MR block. After the $L$ stacked ST-MR blocks, each BSS station gets a learned representation vector that summarizes the spatiotemporal information from multimodal historical demand. Based on the learned representations, we generate future demand predictions using a fully-connected feed-forward network. 

\subsection{Prediction-based Regularization} \label{method:represent_sum}
To train our model more efficiently, we introduce a prediction-based regularization term. Specifically, in addition to bike sharing demand prediction, we generate future predictions of subway and ride-hailing demand with additional feed-forward networks, and use the prediction error of subway and ride-hailing demand as a regularization term in the loss function. The intuition behind it is that by jointly optimizing the demand prediction of auxiliary modes, we can guide the model to extract useful spatiotemporal information from them, which can inturn benefit bike sharing demand prediction. In this way, the loss function is defined as: 
\begin{equation}
\label{eq:loss}
L(\theta) = ||\hat{X}_{b}^{t+1}-{X}_{b}^{t+1}|| + \epsilon_r \left(\sum_{m \in\{s,h\}}{||\hat{X}_{m}^{t+1}-{X}_{m}^{t+1}||}\right)
\end{equation}
where $\hat{X}_{m}^{t+1}, {X}_{m}^{t+1}$ are the predicted and true demand values of mode $m$ at time step $t+1$, $\epsilon_r$ is a pre-determined weight for the prediction-based regularization term.


\section{EXPERIMENTS}

\subsection{Data Description}
Our proposed model is validated on real-world public datasets from NYC. In this study, we use Manhattan as the research area and collect travel demand data of bike sharing, subway and ride-hailing from 2018-03-01 to 2018-08-31. Specifically, we use the following three datasets: (1) \textbf{NYC Citi Bike}: the data consists of the pick-up and drop-off time and station of each Citi Bike trip records. During our study period, there are around 36 thousand trip records per day. We filter out the stations with an average of fewer than three orders per hour and keep 246 BSS stations for demand prediction. (2) \textbf{NYC Subway}: the data contains the entries and exists counts of each turnstile in subway stations every four hours, with around 2.4 million entry/exist counts every day on average. We filter out subway stations with no demand for long time periods, which results in 107 subway stations. (3) \textbf{NYC Ride-hailing}: we use the for-hire vehicle (FHV) data from NYC Taxi \& Limousine Commission (TLC), which is provided by ride-hailing companies such as Uber and Lyft. It contains the pick-up and drop-off time and zone of each trip records during the study period. On average there are 234 thousand trips per day during our study period. The zones are pre-determined by TLC and there are 63 TLC zones in Manhattan.

\subsection{Experiment settings}
The demands of all three modes are aligned into 4-hour intervals and min-max normalization is used for each mode to mitigate the effect of demand variance. Data from the first 60\% time steps are used for model training, the following 20\% for validation, and the last 20\% for model testing. We set the historical time step $T=6$, the number of training epochs $E=500$, the learning rate of 0.002, the batch size of 32 and the dropout ratio of 0.3. To prevent overfitting, we use early stopping on the validation set and a L2 regularization on the loss function with a weight decay of 1e-5. 
The prediction-based regularization weight $\epsilon_r$ is tuned from 0 to 0.3, and we find that our model achieves the best prediction performance for bike sharing demand when $\epsilon_r=0.2$, verifying the effectiveness of jointly optimizing multimodal demand prediction for model training. 

The following models are used as baselines for benchmarking: (1) \textbf{HA}: a statistical method that makes predictions based on the historical average; (2) \textbf{LR}: a regression model that captures the linear relationship between historical and future demand patterns; (3) \textbf{XGBoost}: a gradient boosting machine model to uncover nonlinear relationships; (4) \textbf{LSTM}: a recurrent-based model to capture long- and short-term temporal dependencies in demand sequences; (5) \textbf{STGCN} \cite{yu2017spatio}: a convolutional framework using GCNs for spatial correlations and TCNs for temporal dependencies; (6) \textbf{MGCN} \cite{geng2019spatiotemporal}: a multi-graph convolution network that captures multiple types of spatial correlations with different GCNs; (7) \textbf{Graph WaveNet} \cite{wu2019graph}: a graph learning approach using node embedding to learn an adaptive adjacency matrix. For fair comparison, we use the same experiment settings for all models, and evaluate them using three widely used metrics: Root Mean Square Error (RMSE), Mean Absolute Error (MAE) and Coefficient of Determination ($R^2$). We run 10 independent experiments for each model and report the average values on the test set.

\section{RESULTS}
\subsection{Comparison of Model Performance} \label{res:performance}
\begin{figure}[ht!]
  \centering
  \includegraphics[width=\linewidth]{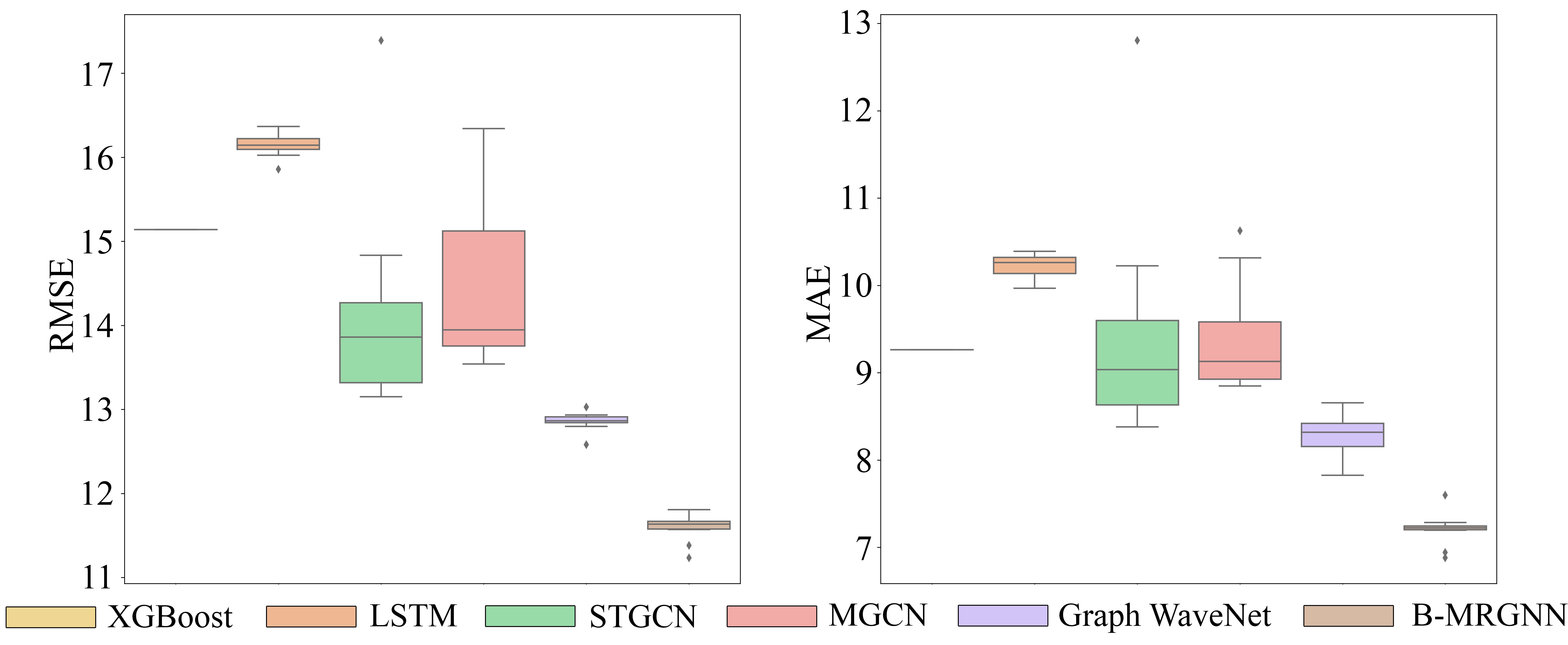}
  \caption{Comparison of model stability}\label{fig:boxplot}
\end{figure}

The performance of different models are summarized in Table~\ref{table:perform_compare}. Compared with the baseline models, our proposed model achieves significantly superior performance for all evaluation metrics. This is likely because our model can directly leverage spatiotemporal information of related subway stations and ride-hailing zones to enhance the prediction of bike sharing demand. In our model, the spatial effects across modes are encoded with inter-modal graphs. 
Figure~\ref{fig:boxplot} presents the performance variance of 10 independent runs of our proposed model and several selected baselines. We can find that our proposed model has the best performance regarding both RMSE and MAE in all experiments with relatively high stability. 

\begin{table}[h]
\caption{Performance comparison of different models}
\label{table:perform_compare}
\begin{center}
\begin{tabular}{|c||c||c||c|}
\hline
Models & RMSE & MAE & $R^2$\\
\hline
HA & 26.832 & 19.322 & 0.167\\
LR & 16.750 & 10.431 & 0.676\\
XGBoost & 15.139 & 9.262 & 0.734\\
LSTM & 16.148 & 10.255 & 0.695\\
STGCN & 13.820 & 9.097 & 0.789\\
MGCN & 14.436 & 9.359 & 0.760\\
Graph WaveNet & 12.689 & 8.092 & 0.814\\
B-MRGNN & \textbf{11.598} & \textbf{7.270} & \textbf{0.845}\\
\hline
\end{tabular}
\end{center}
\end{table}

Among baseline models, the performance of HA is relatively poor, indicating the low temporal regularity of bike sharing demand patterns. This is reasonable as people use bikes for different purposes (e.g. leisure) that are less regular (compared to commuting). LR performs worse than other machine learning models, suggesting the necessity of capturing nonlinear relationships in demand series. XGBoost performs notably better than LSTM, showing the advantage of gradient boosting machines in some cases. The poor performance of LSTM can be potentially explained by its inability to capture spatial dependencies between BSS stations. Although MGCN encodes multiple types of spatial dependencies, it performs worse than STGCN in our experiments, which might be due to its low model stability as illustrated in Figure~\ref{fig:boxplot}. Graph WaveNet achieves the best performance among baseline models given its ability to learn spatial dependencies hidden in demand data with an adaptive adjacency matrix. Compared with Grave WaveNet, our proposed model can further reduce the prediction error, with RMSE and MAE improvement of 8.6\% and 10.2\% respectively.

\subsection{The effect of input modes}

To further investigate the effect of incorporating inter-modal relationships on bike sharing demand prediction, we compare the performance of our model using different input mode combinations: bike sharing alone, bike sharing and subway, bike sharing and ride-hailing, and all three modes. The results are shown in Table~\ref{table:mode_combine}. Note that they share the same model structure, just with different input modes. We find that the inter-modal relations from geographically nearby or semantically similar subway stations and ride-hailing zones can indeed help with the prediction of bike sharing demand: without any inter-modal relations, the prediction error regarding RMSE and MSE would increase by 8.0\% and 8.5\% respectively. Incorporating either subway or ride-hailing demand patterns can already significantly improve the prediction performance of bike sharing demand, with the RMSE reduced by 6.2\% and 7.2\% respectively. 
It is also worth noting that with only bike sharing demand as input, our proposed model can still perform better than Graph WaveNet, suggesting that our spatiotemporal framework can better extract intra-modal relationships. 

\begin{table}[h]
\caption{Performance comparison of different mode combinations}
\label{table:mode_combine}
\begin{center}
\begin{tabular}{|c||c||c||c|}
\hline
Models & RMSE & MAE & $R^2$\\
\hline
Bike sharing & 12.528 & 7.889 & 0.821\\
Bike sharing + subway & 11.754 & 7.391 & 0.843\\
Bike sharing + ride-hailing & 11.630 & 7.275 & \textbf{0.845}\\
Bike sharing + subway + ride-hailing & \textbf{11.598} & \textbf{7.270} & \textbf{0.845}\\
\hline
\end{tabular}
\end{center}
\end{table}

\section{CONCLUSIONS}

This paper proposes a bike sharing demand prediction approach that allows for information sharing across modes. Specifically, we adapt a multi-relational spatiotemporal graph neural network approach to capture the complex spatiotemporal correlations among heterogeneous spatial units from different modes. The spatial dependencies across modes are encoded with multiple intra- and inter-modal graphs, and an inter-modal graph convolution layer is introduced to capture both correlations and difference between nodes from different modes. To integrate dependencies from spatial and temporal domains, we further incorporate MRGNN layers with temporal convolution networks in stacked ST-MR blocks. Empirical validation is performed on real-world bike sharing, subway and ride-hailing datasets from NYC. The results show that our proposed model achieves the best performance compared to existing methods, suggesting that the knowledge of subway and ride-hailing demand can indeed benefit the demand prediction of bike sharing. In future works, this research can be further improved by examining how the effects of cross-mode information vary over space and time.  

\addtolength{\textheight}{-12cm}   







\bibliographystyle{IEEEtran}
\bibliography{ref}

\end{document}